\documentclass[11pt,a4paper]{article}
\usepackage[nohyperref]{acl2018}
\usepackage{times}
\usepackage{latexsym}
\usepackage{amsmath, amsthm, amssymb}
\usepackage{url}
\usepackage{graphicx}
\usepackage{blindtext}
\usepackage{subcaption}
\usepackage{bbm}
\usepackage{etoolbox}
\usepackage{multirow}
\usepackage{multicol}
\usepackage{verbatim}
\usepackage{scalerel,stackengine}
\stackMath
\newcommand\reallywidehat[1]{%
\savestack{\tmpbox}{\stretchto{%
  \scaleto{%
    \scalerel*[\widthof{\ensuremath{#1}}]{\kern-.6pt\bigwedge\kern-.6pt}%
    {\rule[-\textheight/2]{1ex}{\textheight}}
  }{\textheight}%
}{0.5ex}}%
\stackon[1pt]{#1}{\tmpbox}%
}
\makeatletter
\patchcmd\set@fontsize{#4}{#2}{}{}
\makeatother

\aclfinalcopy 

\title{Semi-Supervised Confidence Network aided  Gated Attention based Recurrent Neural Network for Clickbait Detection}

\author{Amrith Rajagopal Setlur \\
  {\tt amrithsetlur22@gmail.com} \\
}

\date{}

\begin{document}
\maketitle
\begin{abstract}

Clickbaits are catchy headlines that are frequently used by social media outlets in order to allure its viewers into clicking them and thus leading them to dubious content. Such venal schemes thrive on exploiting the curiosity of naive social media users, directing traffic to web pages that won't be visited otherwise. In this paper, we propose a novel, semi-supervised classification based approach, that employs attentions sampled from a Gumbel-Softmax distribution to distill contexts that are fairly important in clickbait detection. An additional loss over the attention weights is used to encode prior knowledge. Furthermore, we propose a confidence network that enables learning over weak labels and improves robustness to noisy labels. We show that with merely 30\% of strongly labeled samples we can achieve over 97\% of the accuracy, of current state of the art methods in clickbait detection. 

\end{abstract}

\section{Introduction}

With the number of social media users increasing by the day, one of the prime objectives of online news media agencies is to lead  social media users onto bogus pages through the display of luscious text/images \cite{lowenstein}. In most cases the content on the landing page is disparate to the headline the user clicked on. Source verification is no longer warranted as news agencies aren't held accountable for the content they post online. As \cite{stein} suggests, at least 15 of the most prominent content creators use clickbaits in some form or the other to honey-trap users. Impetus for such schemes can range from directing traffic to web sites that force users to purchase a product, to shaping popular opinion especially during elections. Some clickbaits claim to accomplish inconceivable tasks while others rely on a viewer's inducement to grapevines. 

\begin{itemize}
  \item``You will never believe what this celebrity did at the awards ceremony."
  \item ``10 things that will get you fairer in 5 days."
  \item ``Millionaires want to conceal this scheme. It can make you rich in a week."
\end{itemize}

Earlier approaches on tackling clickbaits mainly focused on cheap and easy to implement solutions. Blacklisting URLs has been, to some extent, effective in regulating an average user's exposure to clickbaits. \cite{gianotto} assumed that most clickbaits are based on a few key phrases, and a naive way yet effective strategy would be to simply look for such phrases. Such an assumption holds no ground today. As the problem grew to be more pervasive, social media companies modeled the probability for a content to be a clickbait based on factors like click-to-share ratio, time spent by user on the target page, and other such quantifiers. Recent research focuses on salvaging sentence structures, n-grams \& embeddings among other features, in classifiers like Random Forests (RF), Gradient Boosted Trees (GBT), Support Vector Machines (SVM) or the vanilla Logistic Regression (LR). With the advent of online news agencies, there exists a plethora of such sources, but labeling each of the headlines from them would be a staggering task. This vindicates the need for an unsupervised / semi-supervised approach.

Contributions of this paper include: 1. A novel loss component on the attention weights that encodes prior information from a weak source of labels, which eventually improves the generalizability of the deep learning model that has been trained on a small representative dataset. 2. A joint architecture that incorporates into the clickbait classification framework a confidence network that tackles label noise. 3. Using Gumbel-Softmax for gated attentions, thus enforcing peaky attentions over word contexts. 4. Empirically proving the performance of the proposed approach on popular clickbait datasets with only a small portion of labeled samples.

\section{Related Work}

The problem of clickbait detection has been primarily studied in two forms. One of them involves a pair of post and target phrases, in which, the objective is to identify whether the post text (visible to the user), is in someway related to the target content (text/images on the landing page). Using this formulation, \cite{biyani} suggested that the features involved in clickbait detection can be broadly classified into: content features (quotes, capitalization), similarity between the source and target representations, informality, forward reference, URLs etc. A gradient boosted tree was trained using these features. \cite{chakra} highlighted the use of features based on linguistic and structural differences between the clickbait \& non-clickbait headlines. Using the n-grams and word patterns, they successfully trained a SVM classifier with a Radial Basis Function (RBF) kernel, that outperformed the baseline rule based method in \cite{gianotto}. 

The other form comprises of making decisions purely based on the headline content \cite{zhou}, \cite{biyani}. Our work is based on this paradigm, and performs on par with methods that follow the former approach. According to \cite{bi_rnn}, Recurrent Neural Network (RNN) based sentence embeddings of the headlines, are expressive enough to learn neural network based classifiers that separate the two classes. \cite{conv_cb} explored convolutional networks that can convolve our word embeddings to learn n-grams, sub-words and token patterns that are consistent with clickbaits. \cite{regres_cb} worked on the twitter dataset and treated clickbait detection as a regression problem instead of binary classification. Hence, they proposed a model that outputs scores indicative of how clickbaity a tweet is. \cite{anand} used a bi-directional LSTM to improve upon the results published by \cite{chakra}, on the \textit{Headlines Dataset} (Section \ref{sec:dataset}).

\cite{attention_sent} employed attention based mechanisms \cite{attention} for text classification. Our work reuses the self-attentive structured sentence embeddings introduced by \cite{bengio}, with some additional components that supplant text classification in semi-supervised environments, with weakly labeled data.

Some of the most recent approaches that have been successful in modelling curiosity are based on the "novelty" and "surprise" components of a headline. \cite{venneti} used topic modelling to identify the topic distributions for each headline in the corpus. Distance metrics in the space of propability distributions like KL divergence and Hellinger distance were used as features to express novelty while surprise was primarily modeled using bi-gram frequency counts. Based on these features, an SVM/LR classifier was trained on a small section of the training data.

\section{Proposed Methodology}

\subsection{Dataset}

\label{sec:dataset}

The first dataset used in the paper is the \textit{Headlines Dataset} curated by \cite{chakra} \footnote{https://github.com/bhargaviparanjape/clickbait/} and used by \cite{anand}, \cite{rony} et al. It holds labels for 32,000 headlines which featured in articles from BuzzFeed, New York Times, Scoopwhoop, Upworthy, The Guardian, ViralNova, The Hindu, ViralStories, Thatscoop and WikiNews. In all, there are 15,999 clickbait and 16,001 non-clickbait samples. We perform an ablation study using a varying proportion of the dataset as our strongly labeled set. Furthermore, we compare our results against strong baselines established on this dataset, in the absence of labeled data. All experiments have been performed using a 5-fold cross validation scheme, in order to reconcile with the existing baselines. 

The second dataset was picked from the \textit{Clickbait-Challenge 2017} \cite{zhou}\footnote{https://www.clickbait-challenge.org/}. The challenge posed clickbait detection as a regression problem, assigning each entry a set of scores from five different annotators. Each $score \in \{0, \frac{1}{3}, \frac{2}{3}, 1\}$\footnote{https://www.clickbait-challenge.org/\#data}. Therefore, each sample was annotated with score summary statistics: “truthMean”, “truthJudgements”, “truthMode”, and “truthMedian”. For the sake of consistency we reformulated this as a classification problem using the following decision:
\begin{equation}
C_{i} = 
\left\{
	\begin{array}{ll}
		clickbait  & \mbox{if } truthMean_{i} \geq 0.5 \\
		non-clickbait  & otherwise
	\end{array}
\right. 
\end{equation}
Although the dataset consists of ``targetText" and ``image" (landing page) data apart from ``postText" (headline), we were able to attain convincing results by using features derived purely from ``postText". This assumption was based on the behaviour of an average annotator, as in  most cases the annotator judgments are purely hinged on the headline. Based on similar assumptions, \cite{anand}, \cite{zhou} and \cite{rony}, used n-grams, simple word filters or latent text representations (LSTMs) that were solely based on headline content. Following 3 splits were provided in the Clickbait-Challenge (C: Clickbaits, NC: Non-clickbaits).

\begin{center}
 \begin{tabular}{|p{1.5cm}|p{1.9cm}|p{1.1 cm}|p{1.1cm}|} 
 \hline
 Label & Headlines & C & NC \\ [0.5ex] 
 \hline
 A & 19,538 & 4,761 & 14,777 \\ 
 \hline
 B & 2,495 & 762 & 1,697 \\ 
 \hline
 C & 80,012 & - & - \\
 \hline
\end{tabular}
\end{center}

We evaluated F1-score and accuracy on the test set B while using portions of the set A as our labeled dataset (with the remaining as unlabeled), bootstrapped with the unlabeled set C.

\subsection{Random Forest}

\label{sec:rf}

In order to identify words of high importance (information gain), we trained a Random Forest Classifier on the strongly labeled section of the dataset. By using entropy (Eq. \ref{eq:ent}) as a measure of information gain, while splitting samples at each node of a tree, we posited that words with low entropy (summarized in Table \ref{tab:rf_table}) were strong signals for identifying clickbaits. Before fitting the random forest, the headlines were pre-processed; all numeric content was mapped to \textless n-token\textgreater, URLs were mapped to \textless u-token\textgreater. Along with these, entity detectors were useful in identifying references to dates/years, locations. The Wordnet Lemmatizer was used to obviate trivial variances in word representations. The analyses presented in Table \ref{tab:rf_table} was done on the \textit{Headlines Dataset}. In this section, we propose the use of Random Forest as a source of weak labels, with the Random Forest being trained on the human labeled samples.

\begin{equation} \label{eq:ent}
\begin{split}
    E_{s}  = \frac{N_{l}}{N} \, E_{l} + \frac{N_{r}}{N} \, E_{r} \\
    E_{l} = - \sum_{i \in C} p_{i_{l}} \log{p_{i_{l}}} \\
    E_{r} = - \sum_{i \in C} p_{i_{r}} \log{p_{i_{r}}} \\ \\
\end{split}
\normalsize
\end{equation}

\begin{center}
\setlength\tabcolsep{0.5pt}
 \begin{tabular}{p{1cm} p{6cm}} 
  $p_{i_{l}}$:&proportion of samples of the left split that $\in$ class $C_i$ \\
  $p_{i_{r}}$:&proportion of samples of the right split that $\in$ class $C_i$ \\
  $N_{l}$:&number of samples in left split \\
  $N_{r}$:&number of samples in right split \\
  $N$:&total number of samples
\end{tabular}
\end{center}

\begin{table}
    \centering
    \begin{tabular}{p{2.2cm} | p{1.6cm} | p{2.2cm}}
         Word & Importance & Naive Inclination \\ [0.5ex] 
         \hline\hline
            \textless n-token\textgreater & 5.120 & clickbait \\
            \hline
            like & 3.112 & clickbait \\ 
            \hline
            dies & 3.042 & non-clickbait \\
            \hline
            people & 2.351 & clickbait \\
            \hline
            know & 2.336 & clickbait \\
            \hline
            life & 2.155 & clickbait \\
            \hline 
            need & 2.062 & clickbait \\
            \hline
            president & 1.799 & non-clickbait \\
            \hline
            wins & 1.664 & non-clickbait \\
            \hline
            kill & 1.623 & non-clickbait \\
            \hline 
            iraq & 1.244 & non-clickbait \\
            \hline
            hilarious & 1.058 & clickbait \\
            \hline
            favorite & 1.039 & clickbait \\
            \hline
            laugh & 0.965 & clickbait \\
            \hline 
            really & 0.963 & clickbait \\
            \hline
            court  & 0.941 & non-clickbait \\
            \hline
            china  & 0.940 & non-clickbait \\
            \hline
            dead &  0.891 & non-clickbait \\
            \hline
            photos & 0.869 & clickbait \\
            \hline
            most & 0.851 & clickbait \\
            \hline 
            leader & 0.702 & clickbait \\
            \hline
            pictures & 0.698 & clickbait \\
            \hline
            obama & 0.592 & non-clickbait \\
            \hline
            questions & 0.524 & clickbait \\
            \hline
    \end{tabular}
    \caption{Some of the tokens with high word-importance factor (Inverse Entropy).}
    \label{tab:rf_table}
\end{table}

Salvaging the tokens with the lowest entropy, we built simple rules to detect clickbaits. Tree paths that included decisions based on these low entropy tokens were used to construct rules in Disjunctive Normal Forms (DNFs) (Table \ref{tab:rules}). The unlabeled dataset was then run through these rules to determine weak labels for classification. This aided the training of the attention network, as corroborated by our experiments (Section \ref{sec:exp}).

\subsection{Problem Definition}

We are given two sets of data, namely: $D_s$\,=\,$\{(X_1, y_1), \, (X_2, y_2), \, .., \, (X_n, y_n)\}$, that is strongly labeled through manual annotations and  $D_w$\,=\,$\{(X_1, {\hat{y}}_1), \, (X_2, {\hat{y}}_2), \, .., \, (X_n, {\hat{y}}_n)\}$, which is weakly labeled and is composed of samples from the unlabeled set. The weak labels in $D_w$ are generated by the Random Forest Classifier (RF) (Section \ref{sec:rf}). In addition to this, we assimilate ${\hat{D}}_s$, that is composed of RF predictions on $D_s$ (Eq. \ref{eq:rfi_s}). 

\begin{equation}
\begin{gathered}
    \label{eq:rfi}
        {\hat{y}}_i = RF(X_i) \\ 
         \forall (X_i, {\hat{y}}_i) \in D_w 
\end{gathered}
\end{equation}
\normalsize

\begin{equation}
\begin{gathered}
    \label{eq:rfi_s}
        {\hat{D}}_s = \{ (X_i, y_i, RF(X_i))\} \\ 
        \forall (X_i, {y}_i) \in D_s
\end{gathered}
\end{equation}

The goal would be to train a classification network on the set obtained by concatenating $D_s$ \& $D_w$. It is assumed that $D_s$ is comprised of a representative set of samples and that $D_s$ and $D_w$ contain i.i.d. samples from the true data distribution.  Since $D_w$ consists of weak labels from the Random Forest and $|D_w| > |D_s|$, we propose the use of a confidence network (Section \ref{sec:opt_alg}) that predicts the accuracy of a weak label. These accuracy scores would re-weigh the gradient updates when the loss is calculated using samples from $D_w$, thus attenuating the effect of noisy labels.

\subsection{Deep Attention Network}

Self-attentive structured attention mechanisms for efficient semantic latent sentence representations was proposed by \cite{smola} and \cite{bengio}. \cite{zhou} further ascertained the effectiveness of the attention mechanism in clickbait detection. The intuition behind a self attention mechanism is that the network learns the importance of each token's context, for the task in hand (clickbait detection), along with a hidden state representation of the word context itself. While \cite{zhou}, \cite{smola} used this in a fully supervised setting, we researched its resilience under the semi-supervised tone. In the following sections, we show that, the presence of an external attention enforcer is pivotal in training the network. Merely training a deep network on a meagre set of a few thousand samples leads to overfitting \cite{dl_book}, and as expected, this claim was substantiated, when we noted a higher validation/test loss, upon direct application of the architecture proposed by \cite{zhou}, \cite{bengio}.

We propose vital and consequential modifications to the base network which accommodates the semi-supervised learning problem. Attention mechanisms have traditionally been utilized to focus attention on features that are most influential in performing a specific task, like image captioning or semantic segmentation, in the image domain \cite{attn_img_cap}. Drawing a parallel, we propose an attention based loss that forces the attentions for a specific set of words to be higher than the rest. When a large number of samples are available, the attention module is self-sufficient in determining such tokens. In this case however, we use as a surrogate, the word importance measures from the Random Forest classifier to identify them (Section \ref{sec:att_loss}).

\subsection{Architecture}
\begin{figure*}[ht]
  \centering
  \includegraphics[width=14cm, height=12cm, keepaspectratio]{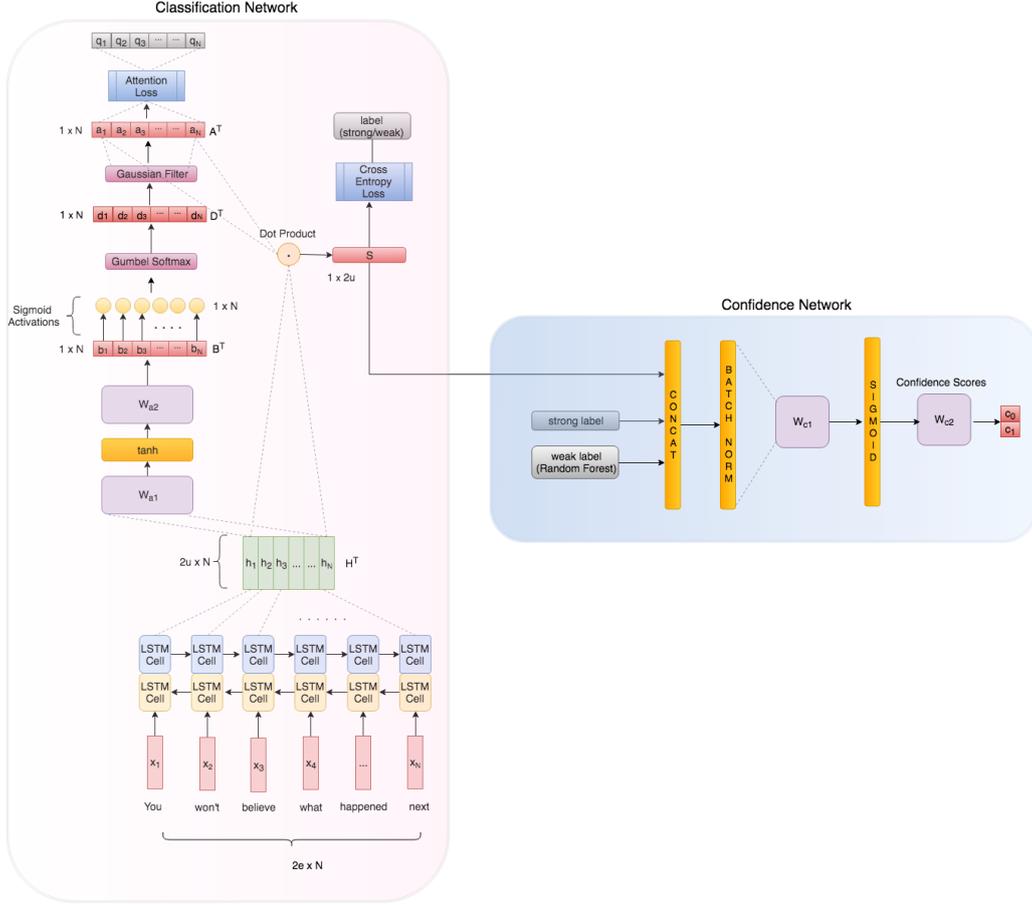}
\caption{Network Architecture}
\label{fig:network}
\end{figure*}
Figure \ref{fig:network} demonstrates the two networks employed during training. The classification/self-attention network determines the sentence embedding post a dot-product operation on the LSTM hidden state representations using the attention weights. Given a headline with $N$ tokens, we map each token $w_i$, where $i \in \{1, \dots, N\}$, to its corresponding glove embedding\footnote{https://nlp.stanford.edu/projects/glove/} (trained on the twitter corpus), denoted by $x_i$. A bi-directional LSTM network, encodes the word context (from both directions), for the word $w_i$, in the time step $t_i$ (Eqs. \ref{eq:lstm_1}).

\begin{equation}
\begin{gathered}
    \label{eq:lstm_1}
        {h_{left}}_i = {LSTM_{left}}(x_i) \\  
        {h_{right}}_i = {LSTM_{right}}(x_i) \\
        {h}_i = {h_{left}}_i \, || \, {h_{right}}_i \\
        x_i \in R^{e}\\
        {h_{left}}_i, {h_{left}}_i \in R^{u}\\
        h_i \in R^{2u}
\end{gathered}
\end{equation}

Given the word context embeddings, the attention network maps them to intermediate token level activations, using a Multi-Layered Perceptron (MLP) network with non-linear (tanh) activations. With $H \in R^{N \times 2u}$ as the context embedding matrix, the network outputs $B \in R^{N}$, as the intermediate activation vector (Eq. \ref{eq:act_vec}).

\begin{equation}
\begin{gathered}
    \label{eq:act_vec}
        B = \sigma((\tanh(H \cdot {W_{a1}}^{T})) \cdot W_{a2})\\
        H \in R^{N \times 2u}, \, W_{a1} \in R^{m \times 2u}, \, W_{a2} \in R^{m \times 1}\\
\end{gathered}
\end{equation}

The original structured self-attentive network \cite{bengio} proposed a softmax activation in the final layer. This was befitting in the case of sentiment classification, where in the sentiment of each sentence was pivoted on a few key tokens and such tokens existed in all sentences, irrespective of the class. On the contrary, in the case of clickbait detection, it is mostly the negative class (clickbaits), that contains such tokens of importance. As depicted in Table \ref{tab:rf_table}, the words identified by the Random Forest for the clickbait class are independent of the news item/news subject (``\textless n-token\textgreater", ``know", ``need", ``favorite", ``most"). Same can't be said for the positive class (``obama", ``iraq", ``china", ``president"). Hence we introduced a sigmoid layer to better suit the case of clickbait detection. 

The intermediate activations $B \in R^N$, obtained post the sigmoid layer were treated as parameters of a binary distribution. Treating each $b_i$ as $P(a_i=1)$, we could sample from the binary distribution. To propagate losses during back-propagation, we instead sampled values from the Gumbel-Softmax distribution \cite{gumbel}, to obtain $D \in R^N$ (Eq. \ref{eq:gumbel}). This encapsulates the ``gated" portion of the network where in activations for word contexts are sampled from a discrete distribution whose parameters are learned. The central idea is to support a case where in the sentence may not have significant impact words (fairly common with the positive class), and with this formulation the sampled values can represent such a case trivially with a zero vector. In the original work done by \cite{bengio}, such activations were inconceivable. 

\small
\begin{equation}
\begin{gathered}
    \label{eq:gumbel}
        d_i = \frac{\exp{((\log(b_i) + g_{i_1})/\tau)}}
    {\exp{((\log(b_i) + g_{i_1})/\tau)} + \exp{((\log(1-b_i) + g_{i_0})/\tau)}}\\
     \\
\end{gathered}
\end{equation}
\normalsize

\begin{center}
\small
\setlength\tabcolsep{0.5pt}
 \begin{tabular}{p{0.4cm} p{6cm}} 
    $g_{i_1}$& $\sim$ Gumbel(0, 1)\\
    $g_{i_0}$& $\sim$ Gumbel(0, 1)\\
    $\tau$:& the temperature parameter that determines 
        the extent to which $d_i$ would be close to 0 or 1
\end{tabular}
\end{center}

Adding the Gumbel-Softmax enables the possibility of peaky activations. Lower values of $\tau$ would lead to $d_i$ lying on either end of the spectrum, i.e 0 or 1, thus resembling a sample from a binomial distribution.

In cases of very low volumes of labeled data we observed that the attention weights of tokens neighboring the words of importance were very close to 0, and thus their context representations wasn't contributing towards the final sentence representation. To avoid this, we used a Gaussian filter over the samples from the Gumbel-Softmax distribution, which acts as a 1D convolution. The weights of the Gaussian filter are not learned. They are fixed and normalized so as to ensure that the final attention weights $0.0 \leq a_i \leq 1.0$. The activation vector $A \in R^N$, and LSTM hidden states $H$ produce the final sentence embedding $S$ (Eq. \ref{eq:sen_emb}).

\begin{equation}
\begin{gathered}
    \label{eq:sen_emb}
        A = GaussianFilter(B)\\
        S = H^\top \cdot A\\
        H \in R^{N \times 2u}, \, A \in R^{N}, \, S \in R^{2u}
\end{gathered}
\end{equation}

\subsection{Attention Loss}

\label{sec:att_loss}

One way to identify key words/features is to use a weak labeler like Random Forest. It identifies a set of words crucial in decision making (Table \ref{tab:rf_table}). The loss penalizing a deviation from such important words, is the standard binary cross entropy loss where it considers the attentions sampled from the Gumbel-Softmax distribution. The true attention is inferred from the word importance or entropy scores generated by the Random Forest. A true attention of 1.0 is assigned to words with importance over a particular threshold. 

The set of positive activations in the corpus is much lower than its counterpart. This class imbalance was tackled by simply using the class weights to re-weigh the attention loss \cite{class_imbalance}. The proportion of activations of class $i$ is inversely proportional to $w_i$ (Eq. \ref{eq:att_loss}). The parameters of the classification network in Figure \ref{fig:network} are optimized over a combination of the attention and classification losses (Eq. \ref{eq:combo_loss}).

\begin{equation}
\begin{gathered}
    \label{eq:att_loss}
        \mathcal{L}_a(X, y) = -W^\top [Q \odot log(A) + \bar{Q} \odot log(\bar{A})] \\
        A = \{a_1,\dots,a_N\}^\top, \, \bar{A} = 1 - A  \\
        Q = \{q_1,\dots,q_N\}^\top, \, \bar{Q} = 1 - Q     \\
        W = \{w_1,\dots,w_N\}^\top
\end{gathered}
\end{equation}

\begin{center}
\setlength\tabcolsep{0.5pt}
 \begin{tabular}{p{1cm} p{6cm}} 
  $A$:&network activations for tokens\\
  $Q$:&true activations for tokens\\
  $W$:&sample weights\\
  $N$:&number of tokens\\
  $\odot$:&Hadamard product\\
  $X, y$:&samples $\sim D_s, D_w$
\end{tabular}
\end{center}

\begin{equation}
\begin{gathered}
    \label{eq:combo_loss}
        \mathcal{L}_{ca}(X, y) = \mathcal{L}_c(X, y) + \lambda \cdot \mathcal{L}_a(X, y)
\end{gathered}
\end{equation}

\begin{center}
\setlength\tabcolsep{0.5pt}
 \begin{tabular}{p{1cm} p{6cm}} 
  $X, y$:&samples $\sim D_s, D_w$\\
  $\mathcal{L}_{ca}$:&Classification + Attention Loss\\
  $\mathcal{L}_c$:&Classification Loss\\
  $\mathcal{L}_a$:&Attention Loss\\
  $\lambda$:&Contribution of the attention loss to the total loss 
\end{tabular}
\end{center}


\subsection{Optimization Algorithm}
\label{sec:opt_alg}

Amalgamation of strong and weak supervision has been used to solve constraints like data paucity, noisy labels and a few others \cite{weak_learners}. Weak learners are sources of noisy labels. \cite{deghani} proposed the use of a confidence network to estimate the accuracy of a noisy label. Similar to \cite{deghani_fidelity} and \cite{arachie}, we used an optimization method that is a variant of the standard Stochastic Gradient Descent (SGD). The latter uses a constant learning rate on all samples in an iteration. Such an approach can lead to noisy gradients, especially when the proportion of strongly labeled samples in a batch is low \cite{dl_book}. 

In order to mitigate this, a confidence network, trained on ${\hat{D}}_s$, formulates the confidence in a weak label as a function of the weak label (Random Classifier output) and the encoded input representation. As shown in Figure \ref{fig:network}, the architecture is split into the classification and confidence networks. Batches of i.i.d samples are drawn iteratively from ${\hat{D}_s}$ and $D_w$. A batch of the former type is used train the classification network, the weights for which are updated using SGD. Since the samples from ${\hat{D}_s}$ also consist of RF predictions on strongly labeled samples, they are used to train the confidence network ($f_{conf}$) as well.  

\begin{equation}
\begin{gathered}
    \label{eq:conf_sc}
        score = f_{conf}(E(X), {\hat{y}}) \\ 
         \forall (X, y, {\hat{y}}) \in {\hat{D}}_s, \, \\
         E(X): latent \, representation \, for \, X
\end{gathered}
\end{equation}

For a sample $(X, y, \hat{y}) \sim {\hat{D}}_s$, a forward propagation on the classification network would generate the sentence representation $S$ (denoted as $E(X)$ in Eq. \ref{eq:conf_sc}). $S$ is concatenated with $\hat{y}$, and passed to a batch-normalization layer. Since the embeddings and binary signals lie on separate manifolds, the batch-normalization is quint-essential before the concatenated input is passed to a neural network. Figure \ref{fig:network} enunciates the output of the network as a 2-dimensional vector $[c_0, c_1]$. The $score$ in Eq. \ref{eq:conf_sc} is given by $c_1$. The confidence network is trained using a cross-entropy loss ($\mathcal{L}_{conf}$ in Eq. \ref{eq:conf_nw_loss}). The true confidence is given by the indicator  $\mathbbm{1}_{y=\hat{y}}$. 

In the subsequent iteration when samples are drawn from $D_w$ confidence scores are inferred on each sample in this set through a forward propagation on the confidence network. These scores are passed to the optimization algorithm for the classification network in order for it to re-weigh its gradient updates. Equation \ref{eq:grad_upd} defines the gradient update rule for parameters in the classification network.

\begin{equation}
\begin{gathered}
    \label{eq:conf_nw_loss}
        \mathcal{L}_{conf}(X, y, \hat{y}) =  - \mathbbm{1}_{y\neq\hat{y}} \log{c_0}  - \mathbbm{1}_{y=\hat{y}} \log{c_1}\\
\end{gathered}
\end{equation}

\begin{center}
 \begin{tabular}{p{7cm}} 
  $(X, y, \hat{y}) \sim {\hat{D}}_s$ \\
  $c$ : $[c_0, c_1]$, output of the confidence network \\
  $\mathcal{L}_{conf}$ : Loss on sample $(X, y, \hat{y})$\\
  $\mathbbm{1}_{y=\hat{y}}$ : True confidence value
\end{tabular}
\end{center}

\begin{equation}
\begin{gathered}
    \label{eq:grad_upd}
        \reallywidehat{\nabla w} = \frac{1}{B}\sum_{i=1}^{B}{f_{conf}(X_i, \hat{y}_i)}\cdot \nabla_{w}\mathcal{L}_{ca}(X_i, \hat{y}_i) \\
        w_{t+1} =  w_{t} - \eta_{t}\reallywidehat{\nabla w}\\
\end{gathered}
\end{equation}

\begin{center}
\setlength\tabcolsep{0.5pt}
 \begin{tabular}{p{1cm} p{6cm}} 
  $\eta_{t}$:&learning rate \\
  $B$:&Batch size of samples drawn from ${\hat{D}}_s$\\ 
  $\mathcal{L}_{ca}$:&Loss on sample $(X_i, \hat{y}_i)$ (Section \ref{sec:att_loss})\\
  $f_{conf}$:&Confidence network
\end{tabular}
\end{center}

\section{Experiments}

\label{sec:exp}

\subsection{Experimental Setting}

The following discussion on hyper-parameters is with respect to the \textit{Headlines Dataset}. Although nearly the same set of values were applicable to the \textit{Clickbait-Challenge Dataset}.  

The hyper-parameters of the random forest were fine tuned using Bayesian optimization techniques. We used 50 estimators, with a maximum depth of 3 and a minimum split size of 5 along with the entropy criterion for splitting. Rules in DNF form were constructed by traversing paths that consisted only of tokens with high information gain. The threshold for minimum word importance was found to be optimal at 0.42. Table \ref{tab:rules} lists some of the rules (mentioning only the words whose presence triggers the corresponding rule).

\begin{table}
    \centering
    \begin{tabular}{|p{5.3cm} | p{1.0cm}|} 
    \hline
      \textbf{Rule} & \textbf{Class} \\
      \hline
      believe $\land$ \textless n-token\textgreater & C \\
      \hline
      president $\land$ iraq $\land$ war & NC \\
      \hline
      hilarious $\land$ photos $\land$ \textless n-token\textgreater & C \\
      \hline
     \end{tabular}
    \caption{Rules drawn from the tress in RF.}
    \label{tab:rules}
\end{table}

The dimensionality of the parameters entrenched in the classification and confidence networks have been encapsulated in Table \ref{tab:hyp}. Glove embeddings \cite{glove} trained on the twitter corpus, were used as base word embeddings, which were fed to the LSTM cells. Dropouts \cite{dropout} have been commonly used in recurrent neural networks as a form of regularizer, especially after they were proven to be a form of Bayesian inference in deep Gaussian Processes \cite{dropoutbayes}. We used a dropout parameter of 0.5 within the LSTM, and 0.4 for the inputs to the fully connected layer with weights $W_{a2}$. For the Gumbel-Softmax layer, temperature parameter of $\tau=0.7$ was found to be optimal. Lower values of $\tau$ would have led to higher attention weights for the words of importance, but it would have also driven the weights for the rest of the tokens to zero. For the batch-normalization layer present within the confidence network, a momentum of 0.05 resulted in a slightly better validation accuracy as compared to the standard value of 0.1 \cite{batch_norm}. This can be attributed to the variance in label confidence across alternate mini-batches that were sampled randomly from the labeled and unlabeled sets. The standard mini-batch SGD optimizer with a learning rate of 0.0001 \cite{sgd} and a batch size of 64 samples was employed. The parameter $\lambda$, involved in the combination of attention and classification losses was fixed at 0.3. We used an early stopping criteria, to stunt training. In most cases, 5 epochs were sufficient to fit the training data, a result, which seems to corroborate with the size of the dataset involved.

\begin{table}
    \centering
    \begin{tabular}{|p{3.75cm} | p{2.5cm}|}
        \hline
         \textbf{Parameter} & \textbf{Dimensionality} \\ 
         \hline
         $x$  (Word Embedding) & $300$ \\
         \hline
         $h$  (LSTM hidden state) & $200$ \\
         \hline
         $W_{a1}$ & $32 \times 400$\\
         \hline
         $W_{a2}$ & $32 \times 1$\\
         \hline
         $W_{c1}$ & $64 \times 400$\\
         \hline
         $W_{c2}\footnotemark$ & $65 \times 2$ \\
         \hline
    \end{tabular}
    \caption{Dimensionality of the parameters in the classification and confidence networks.}
    \label{tab:hyp}
\end{table}
\footnotetext{This includes the bias terms as well.}

\subsection{Results}

In accordance with the disparity of the datasets mentioned in section \ref{sec:dataset}, we summarize the proposed model performance on the two of them independently. In both cases, we benchmark the model performance against baselines, that claim to have achieved state of the art results on the dataset in question.

\cite{anand} claimed a high classification accuracy of 0.982 after having experimented with multiple RNN based architectures to embed the clickbait embeddings in a multi-dimensional space. Our model's performance emulates the former on the fully labeled dataset. On the partially labeled set we achieve a high accuracy of 0.980, even with only 30\% of labeled samples (Table \ref{tab:baseline_1}). This is an increment of \textbf{4.03\%} in the accuracy on the validation set, when compared to the BiLSTM based network.

\begin{table*}[ht]
    \centering
        \begin{tabular}{|p{5.8cm}|p{1.8cm}|p{1.8cm}|p{1.8cm}|p{1.8cm}|}
            \hline
            Model & Accuracy & Precision & Recall & ROC-AUC \\
            \hline
    Baseline (BiLSTMs)                   & 0.982 & \textbf{0.984} & 0.980 & \textbf{0.998} \\ 
    Self-Attentive Network (SAN)         & 0.982 & 0.983 & 0.981 & 0.997 \\
    SAN + Gumbel Softmax (GS)            & 0.981 & 0.982 & 0.981 & 0.996 \\
    SAN + GS + Gaussian Filter (GF)      & \textbf{0.983} & \textbf{0.984} & \textbf{0.982} & 0.997 \\
        \hline
        \end{tabular}
    \caption{Performance metrics against the existing baseline \cite{anand}, with the fully labeled \textit{Headlines Dataset}.  [Averaged over a 5-fold cross validation scheme]}
    \label{tab:baseline_1}
\end{table*}

\begin{table*}[ht]
    \centering
        \begin{tabular}{|p{4cm}|p{0.78cm}|p{0.78cm}|p{0.78cm}|p{0.78cm}|p{0.78cm}|p{0.78cm}|p{0.78cm}|p{0.78cm}|p{0.78cm}|}
            \hline
            \multirow{2}{*}{Model} & \multicolumn{3}{|c|}{Accuracy} & \multicolumn{3}{|c|}{Precision} & \multicolumn{3}{|c|}{Recall} \\
                                   \cline{2-10}
                                   &    80\% & 50\% & 30\% & 80\% & 50\% & 30\% & 80\% & 50\% & 30\%  \\  
            
            \hline
            Baseline (BiLSTMs)           &  0.980 & 0.966 & \textbf{0.942} & 0.979 & 0.967 & \textbf{0.943} & 0.981 & 0.966 & \textbf{0.944}       \\
            SAN                          &  0.979 & 0.966 & 0.944 & 0.978 & 0.965 & 0.945 & 0.980 & 0.967 & 0.942       \\
    SAN + RF                             &  0.980 & 0.976 & 0.959 & 0.980 & 0.974 & 0.958 & 0.981 & 0.976 & 0.960       \\
    SAN + RF + GS                        &  0.980 & 0.978 & 0.970 & 0.981 & 0.978 & 0.971 & 0.982 & 0.979 & 0.972       \\
    SAN + RF + GS + GF                   &  0.981 & 0.977 & 0.975 & 0.982 & 0.978 & 0.976 & 0.981 & 0.977 & 0.977       \\
    SAN + RF + GS + GF + Confidence N/w  &  0.983 & 0.982 & \textbf{0.980} & 0.984 & 0.983 & \textbf{0.980} & 0.982 & 0.982 & \textbf{0.981}       \\
    \hline
        \end{tabular}
    \caption{Ablation study with variations in the proportions (80\%, 50\%, \& 30\%) of labeled  data (\textit{Headlines Dataset}). [Averaged over a 5-fold cross validation scheme]\\}
    \label{tab:baseline_1_ablation}

        \centering
        \begin{tabular}{|p{3cm}|p{0.85cm}|p{0.85cm}|p{0.85cm}|p{0.85cm}|p{0.85cm}|p{0.85cm}|p{0.85cm}|p{0.85cm}|p{0.85cm}|}
            \hline
            \multirow{2}{*}{Model} & \multicolumn{3}{|c|}{MSE} & \multicolumn{3}{|c|}{F1-Score} & \multicolumn{3}{|c|}{Accuracy} \\
                                   \cline{2-10}
                                   &    100\% & 50\% & 30\% & 100\% & 50\% & 30\% & 100\% & 50\% & 30\%  \\  
            
            \hline
             Baseline & \textbf{0.033}  & 0.047 & 0.055 & \textbf{0.683} & 0.557 & 0.521 & \textbf{0.856} & 0.713 & 0.671  \\
             \hline
             Proposed Model  &  0.034 & \textbf{0.038} & \textbf{0.042} & 0.679 & \textbf{0.668} & \textbf{0.662}  & \textbf{0.856} & \textbf{0.835} & \textbf{0.829}       \\
    \hline
        \end{tabular}
    \caption{MSE, F1-Score and Accuracy metrics for the existing baseline \cite{zhou} \& our solution on the test set, while using different proportions of set A (100\%, 50\% \& 30\%) as our labeled data.}
    \label{tab:baseline_2}
    
\end{table*}

We further study model performances across the various model architectures supplanted in an incremental fashion (Table \ref{tab:baseline_1_ablation}). When data paucity is high (30\% labeled), we see significant differences in accuracy, precision and recall while adding  individual components to the network. Increments have been noticed when a large number of labeled samples (80\%) are available, albeit the rate of improvement is negligible. The gaussian filter over attention weights sampled from Gumbel-Softmax is more effective in the former case where scattering attention onto the neighborhood of high importance words prevents the sentence representation from collapsing to an average of word vectors in the inchoate stages of training. The precision and recall metrics also show similar trends, with increments of \textbf{3.92\%} and \textbf{3.91\%} respectively, in case of 30\% labeled samples.

In case of the \textit{Clickbait-Challenge} dataset (section \ref{sec:dataset}), we train our model on the labeled split (A), concatenated with the unlabeled samples in (C) and observe model performance on the test set (B),  consistent with \cite{zhou}. We further performed experiments to study our model's ability to learn meaningful sentence representations when only a portion of the set A was labeled. Table \ref{tab:baseline_2} puts into perspective the observed Mean Squared Error (MSE), precision and accuracy on the test set, in comparison to the current best performing model \cite{zhou} on the dataset in question. The existing baseline involves sentence classification solely using the self-attention network introduced by \cite{bengio}. The baseline results are convincing when all labels are available and on par with our model's performance. On the other hand, when only 30\% of the set A is used as the labeled set we observed a jump of \textbf{23.55\%} \& \textbf{25.61\%} with regards to the accuracy \& F1-score respectively.

\section*{Conclusion and Future Work}

In this paper, we proposed a novel architecture to tackle clickbait detection when only a few labeled samples are present. We successfully showed that use of a weak labeler like Random Forest can generate priors over an attention mechanism and thus improve generalizability. Empirically, we have also shown that training a confidence network to rescale gradients, helps tackle the inherent noise attributed to the presence of a weak labeler. 

We haven't considered $\lambda$ as a function of time, and annealing $\lambda$ over time may improve performance. Future work may also involve confidence networks with, momentum and adaptive learning rate based gradient update methods like Adam or RMSprop \cite{adam}. The glove embeddings capture the word contexts. Modelling curiosity \cite{venneti} in conjunction with such features may help capture user intent as well. Nevertheless, this requires a different set of experiments and benchmarks to thoroughly understand the intricacies involved in such a mixture. 

\bibliography{acl2018}
\bibliographystyle{acl_natbib}

\end{document}